 \documentclass[smallabstract,smallcaptions]{dccpaper}

\usepackage{cite}
\usepackage{graphicx}
\usepackage{url}
\usepackage{amsmath}
\usepackage{amssymb}
\usepackage[caption = false, font=footnotesize]{subfig}
\usepackage{float}
\usepackage{booktabs} 
\usepackage{multirow}
\usepackage[colorlinks=true,linkcolor=blue]{hyperref}%
\usepackage{array}
\usepackage{color}
\usepackage{tabularx}
\usepackage{longtable}
\usepackage{tabu}
\usepackage{pifont}
\usepackage{amssymb}
\usepackage{ragged2e}
\usepackage{tabularx}
\usepackage{longtable}
\usepackage{booktabs}
\usepackage{rotating}
\usepackage{makecell}
\begin{document}

\newcommand\blfootnote[1]{%
  \begingroup
  \renewcommand\thefootnote{}\footnote{#1}%
  \addtocounter{footnote}{-1}%
  \endgroup
}

\title{\textbf{Sparse2Dense: A Keypoint-driven Generative Framework for Human Video Compression and Vertex Prediction}}

\author{Bolin Chen$^{\natural }$$^{\dag}$$^{\emptyset}$, Ru-Ling Liao$^{\dag}$, Yan Ye$^{\dag}$, Jie Chen$^{\dag}$$^{\emptyset}$, \\ Shanzhi Yin$^{\ast}$, Xinrui Ju$^{\ast}$, Shiqi Wang$^{\ast}$ and Yibo Fan$^{\natural }$  \\[0.5em]
{\small\begin{minipage}{\linewidth}\begin{center}
\begin{tabular}{c}
$^{\natural}$ Fudan University \\
$^{\dag}$ DAMO Academy, Alibaba Group \\
$^{\emptyset}$ Hupan Lab \\
$^{\ast}$ City University of Hong Kong
\end{tabular}
\end{center}\end{minipage}}}

\maketitle
\begin{abstract}
For bandwidth-constrained multimedia applications, simultaneously achieving ultra-low bitrate human video compression and accurate vertex prediction remains a critical challenge, as it demands the harmonization of dynamic motion modeling, detailed appearance synthesis, and geometric consistency. To address this challenge, we propose Sparse2Dense, a keypoint-driven generative framework that leverages extremely sparse 3D keypoints as compact transmitted symbols to enable ultra-low bitrate human video compression and precise human vertex prediction. The key innovation is the multi-task learning-based and keypoint-aware deep generative model, which could encode complex human motion via compact 3D keypoints and leverage these sparse keypoints to estimate dense motion for video synthesis with temporal coherence and realistic textures. Additionally, a vertex predictor is integrated to learn human vertex geometry through joint optimization with video generation, ensuring alignment between visual content and geometric structure. Extensive experiments demonstrate that the proposed Sparse2Dense framework achieves competitive compression performance for human video over traditional/generative video codecs, whilst enabling precise human vertex prediction for downstream geometry applications. As such, Sparse2Dense is expected to facilitate bandwidth-efficient human-centric media transmission, such as real-time motion analysis, virtual human animation, and immersive entertainment.
\end{abstract}

\vspace{-0.6em}
\section{Introduction}
\vspace{-0.6em}

Human-centric media transmission in multimedia systems faces critical challenges in delivering ultra-low bitrate high-quality human video streams. This stems from the complex interplay of dynamic motion, photorealistic fidelity, and precise geometry modeling. Efficient compression reduces bandwidth demands through data optimization, while vertex prediction lowers geometric complexity via mesh deformation. These technologies jointly address the quality-efficiency trade-off in real-time VR/AR, telepresence, and immersive applications.
In particular, traditional video compression standards, such as H.265/High Efficiency Video Coding (HEVC)~\cite{sullivan2012overview} and H.266/Versatile Video Coding (VVC)~\cite{bross2021overview}, prioritize general-purpose efficiency but often struggle to exploit the prior knowledge or structural statistics of human motion and geometry, resulting in suboptimal performance for human-centric content. 
While recent advances in generative models (e.g., Generative Adversarial Nets (GANs)~\cite{goodfellow2014generative} and Diffusion Models (DMs)~\cite{dhariwal2021diffusion}) have enabled high-fidelity video synthesis~\cite{liu2021generative,blattmann2023align} and animation~\cite{wang2021one,xu2024magicanimate}, these traditional methods require high-dimensional pixel-space instead of compact latent representations for vivid signal reconstruction, limiting their applicability to bandwidth-constrained scenarios. 

Recently, Generative Video Compression (GVC) algorithms~\cite{chen2024generative,chen2024generative2,chen2025generative,11002361,10109861,chen2024standardizing} have begun to leverage the powerful capabilities of generative models to revive traditional Model-Based Coding (MBC) paradigms~\cite{pearson1995developments,aizawa1995model}, achieving high-quality video reconstruction via highly-compact representations like 2D/3D keypoints~\cite{9859831}, implicit feature~\cite{chen2025rethinking}, motion vectors~\cite{yin2024generative}, human semantics~\cite{chen2025compressing}. However, these approaches primarily focus on photorealistic 2D signal reconstruction, lacking explicit integration with 3D geometric modeling. In other words, most frameworks treat video compression and vertex prediction~\cite{lin2023one,zhang2019predicting,chun2023representation} as separate tasks, neglecting the intrinsic correlation between visual content and geometric structure. This disjointed approach hinders the development of unified solutions for applications requiring both high-quality video reconstruction and accurate human modeling. To further facilitate such application possibilities, Chen \textit{et al.}~\cite{chen2025compressing} introduced a framework for compressing human body videos by integrating interactive semantics via the Skinned Multi-Person Linear model (SMPL)~\cite{SMPL2015}, enabling video coding to achieve interactivity and user-controllability through manipulation of semantic-level representations and reconstruction of 3D human vertex. However, this approach requires integrating an external SMPL model for inference, which increases the overall system complexity.

\begin{figure*}[t]
\centering
\vspace{-0.8cm}
\centerline{\includegraphics[width=1 \textwidth]{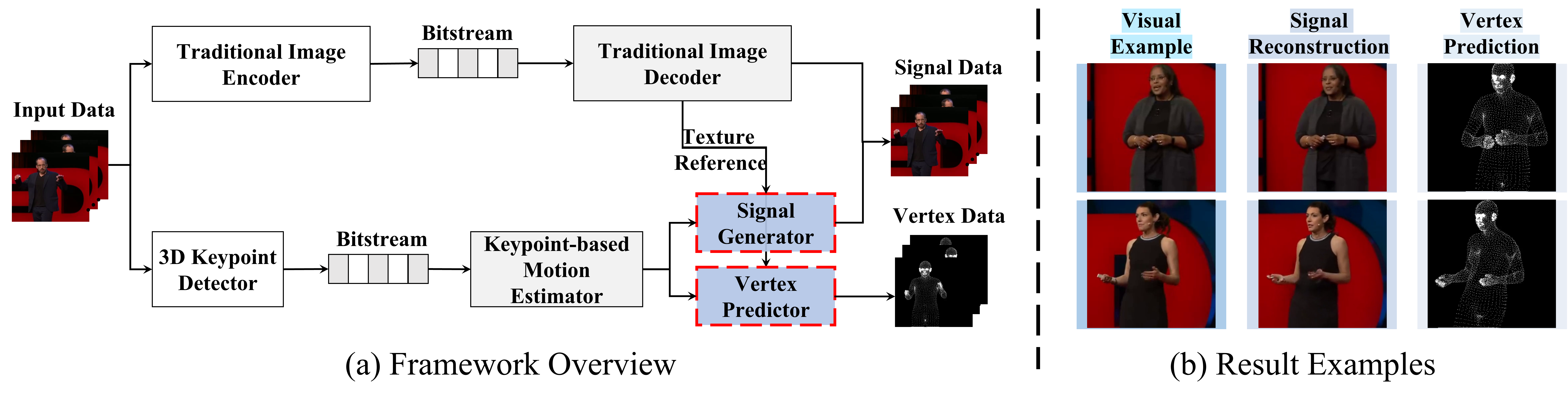}}
\vspace{-0.5cm}
\caption{High-quality signal reconstruction and vertex prediction achieved using our proposed Sparse2Dense algorithm.} 
\vspace{-0.5em}
\label{fig_exmaples}
\end{figure*}

As such, a critical bottleneck in this domain lies in the trade-off between compression efficiency, reconstruction quality, and geometric accuracy within a generative human video compression framework that enables versatile tasks. As illustrated in Figure~\ref{fig_exmaples}, we propose Sparse2Dense, a novel keypoint-driven generative framework that simultaneously achieves ultra-low bitrate human video compression and precise vertex/geometry prediction. 
Our method leverages these extremely sparse 3D keypoints as the compact transmitted symbols, enabling efficient encoding of complex human motion while preserving rich semantic information for signal reconstruction and vertex prediction. 
The core innovation lies in a multi-task learning-based and keypoint-aware deep generative model that could simulate motion dynamics from compact keypoint representations to reconstruct realistic videos and predict human vertex.
This end-to-end framework bridges the gap between low-bitrate compression and high-fidelity geometric modeling, offering a unified solution for bandwidth-efficient human-centric media transmission. The main contributions of this paper are summarized as follows,
\begin{itemize}
\vspace{-0.2cm}
\item{\textbf{Unified Framework:} We propose Sparse2Dense, a novel framework to jointly optimize ultra-low bitrate human video compression and vertex prediction from compact representation, bridging the gap between visual fidelity and geometric accuracy. As such, the proposed Sparse2Dense is expected to enable bandwidth-efficient solutions for human video communication and analysis.} 
\vspace{-0.2cm}
\item{\textbf{Keypoint-driven Compression:} We design a keypoint-aware generative model that can characterize complex human motion with a series of compact 3D keypoints and facilitate the evolution of dense motion fields from these 3D sparse keypoints for high-quality synthesis, thus achieving competitive compression gains especially at ultra-low bitrate ranges over conventional video codecs.} 
\vspace{-0.2cm}
\item{\textbf{Vertex-aware Prediction:} We develop a learnable vertex predictor integrated with the frame generation module, ensuring alignment between visual content and geometry structure through joint optimization. As such, this module could enhance signal synthesis fidelity and predict explicit vertex information for downstream applications like motion analysis.} 
 
\end{itemize}

\begin{figure*}[t]
\centering
\vspace{-1.5cm}
\centerline{\includegraphics[width=1 \textwidth]{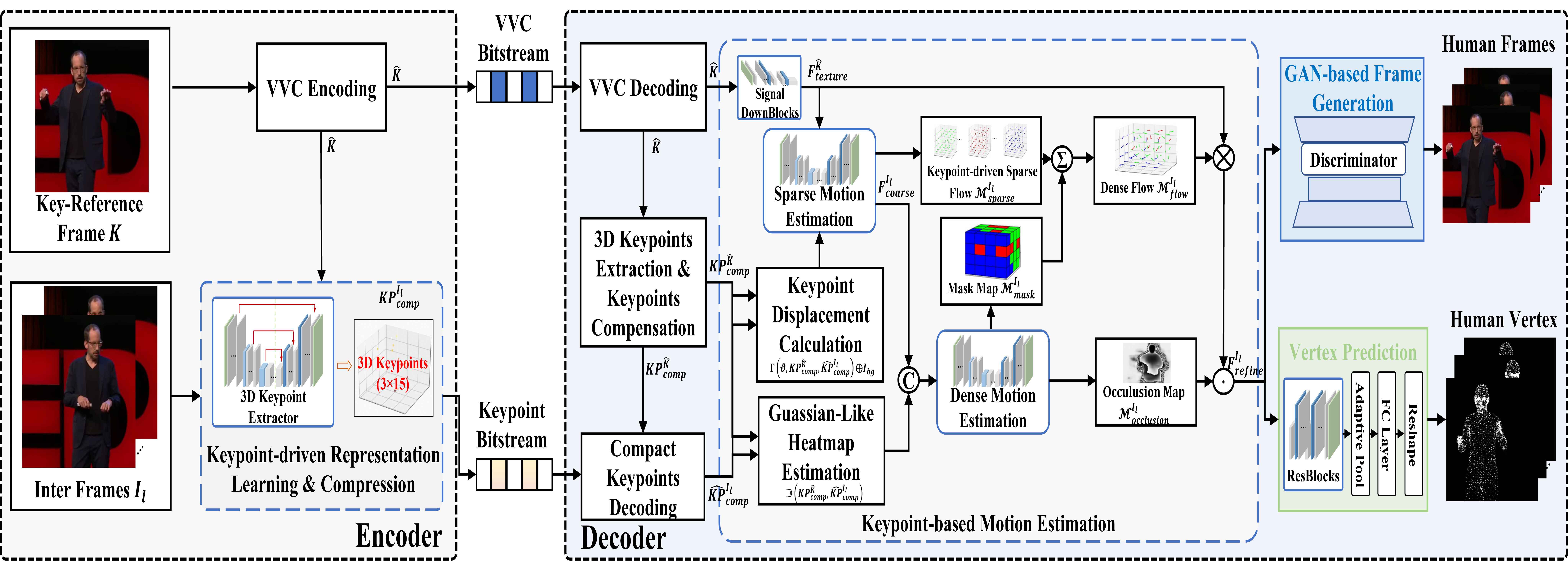}}
\caption{Overview of generative human video coding framework with the proposed Sparse2Dense scheme towards high-efficiency human body video reconstruction and vertex prediction.}
\vspace{-1em}
\label{framework}
\end{figure*}

\vspace{-2mm}
\section{The Proposed Sparse2Dense Framework}
The proposed Sparse2Dense is a keypoint-driven generative framework designed to address the dual challenges of ultra-low bitrate human video compression and precise 3D vertex prediction under bandwidth constraints. The framework operates through a multi-task learning-based, keypoint-aware architecture that harmonizes dynamic motion modeling, detailed appearance synthesis, and geometric consistency. As illustrated in Figure~\ref{framework}, the encoding process begins with compressing the key-reference frame containing human body textures using the state-of-the-art VVC codec, which serves as a texture reference for signal reconstruction. Afterwards, the subsequent inter frames are then processed by a keypoint-driven neural prediction network to generate 15 sets of 3D keypoints. Finally, these highly-sparse keypoints encompassing both spatial and temporal information undergo temporal prediction, quantization, and entropy encoding to produce the final transmitted bitstream. 

Upon receiving the bitstream, the decoder side executes motion field estimation, signal reconstruction, and vertex prediction. Initially, the key-reference frame is decoded using a VVC-based decoder and transformed into 3D keypoints. Additionally, the compact keypoints from inter frames are recovered through entropy decoding combined with temporal compensation. Afterwards, the 3D keypoint representations from both key-reference and inter frames are utilized to compute dense motion fields (\textit{i.e.,} optical flow and occlusion maps). Finally, the decoded key-reference frame and motion fields are fed into a hybrid multi-task neural network architecture containing separate branches for signal synthesis and vertex prediction, enabling the generation of high-fidelity human body signal alongside precise human vertex.

\subsection{Keypoint-driven Representation Learning \& Compression} 
Analogous to FV2V~\cite{wang2021one}, high-dimensional signals—such as the VVC reconstructed key-reference frame $\hat{K}$ or the subsequent inter frames $I_{l} \left (1\le l \le n , l\in Z   \right ) $ can be encoded into temporally compact 3D keypoint representations in a self-supervised learning manner.
This process is structured as three sequential stages: (1) spatial downsampling, (2) U-Net heatmap learning, and (3) 3D keypoint evolution.
Firstly, the input frames $\hat{K}$ and $I_{l}$ undergo spatial downsampling with a scaling factor $s$ to mitigate spatial redundancy and accelerate training efficiency. Next, the reduced-resolution frames are fed into a U-Net architecture~\cite{RFB15a}, which generates hierarchical heatmap representations by integrating encoder-decoder structures with skip connections. Finally, these heatmap representations are further transformed into 15 sets of 3D keypoints by projecting them onto a 3D coordinate grid and spatially summing the grid-based heatmap values. These transformed 3D keypoints to be compressed $ KP_{comp}$ can be descried as follows,
\begin{equation}
 KP_{comp}= \varrho_{\left (grid  \right )}\left ( f_{U-Net}\left ( \phi \left ( X,s \right ) \right ) \right ),
\end{equation}  where $\phi\left ( \cdot \right )$, $f_{U-Net}\left ( \cdot \right )$ and $\varrho_{\left (grid  \right )} \left ( \cdot \right )$ are down-sample operation, U-Net heatmap learning and keypoint transformation with gird-based heatmap summation, respectively. Herein, $X$ could be $\hat{K}$ or $I_{l}$.

To improve the compression efficiency of  compact 3D keypoints, a temporal prediction mechanism is implemented by performing inter-frame prediction between the current frame’s keypoints with those of the previously coded frame. 
The obtained inter-frame keypoint residuals are then quantized and encoded via context-adaptive entropy coding~\cite{TEUHOLA1978308,1096090}, which adaptively optimizes probability distributions according to contextual information to enhance bitstream efficiency.

\vspace{-0.3cm}
\subsection{Keypoint-based Motion Estimation}
\vspace{-0.2cm}
To improve motion estimation accuracy in dynamic human body modeling, we employ a keypoint-aware motion estimation module based on a sparse-to-dense strategy~\cite{Siarohin_2019_NeurIPS,wang2021one}. As such, our method can transform compact 3D keypoint representations into dense motion fields for robust signal reconstruction.

In the decoding phase, the key-reference frame $\hat{K}$ undergoes VVC-based decoding followed by two essential stages: (1) texture feature transformation, where the decoded frame is passed through the feature learning network to generate texture feature $F_{texture}^{\hat{K}}$, and (2) 3D keypoints localization, where composite 3D keypoints $KP^{\hat{K}}_{comp}$ are obtained via a specialized keypoint extraction network. In addition, the 3D keypoints of inter frames $\hat{KP}^{I_{l}}_{comp}$ are reconstructed via arithmetic decoding combined with feature compensation mechanisms.

Subsequently, a sparse motion field $\mathcal M_{sparse}^{I_{l}}$ is constructed by initializing a 3D coordinate grid $\vartheta$, computing displacement vectors between $KP^{\hat{K}}_{comp}$ and $\hat{KP}^{I_{l}}_{comp}$, and integrating background preservation $I_{bg}$ to ensure spatial coherence in dynamic transformations. Moreover, a warping operation $f_{w}$ applies the sparse motion field $\mathcal M_{sparse}^{I_{l}}$ to texture feature $F_{texture}^{\hat{K}}$, producing coarse deformed feature $F_{coarse}^{I_{l}}$ that capture global pose variations. The overall process can be formulated as,
\begin{equation}
 F_{coarse}^{I_{l}}= f_{w}\left (F_{texture}^{\hat{K}}, \Gamma \left (\vartheta, KP^{\hat{K}}_{comp},\hat{KP}^{I_{l}}_{comp}   \right ) \oplus I_{bg} \right ), 
\end{equation} where $\Gamma\left ( \cdot \right )$ and $\oplus$ represent a transformation operator combining displacement computation, and a concatenation operator to integrate background and keypoint-driven motions, respectively.

In addition, $KP^{\hat{K}}_{comp}$ and $\hat{KP}^{I_{l}}_{comp}$ are converted into the corresponding Gaussian-like heatmaps via Gaussian-like function~\cite{Siarohin_2019_NeurIPS}, and the heatmap difference is further computed to form a spatial guidance tensor for deformation modeling. Such spatial heatmap difference is concatenated with $F_{coarse}^{I_{l}}$ and further fed into a U-Net architecture to output a mask map $\mathcal M^{I_{l}}_{mask}$ and an occlusion map $\mathcal M^{I_{l}}_{occlusion}$. In particular, the mask map is utilized to refine the sparse motion field $\mathcal M_{sparse}^{I_{l}}$ to produce the dense motion flow $\mathcal M^{I_{l}}_{flow}$. The detailed process can be represented as follows,
\begin{equation}
\mathcal M^{I_{l}}_{flow}= P_{1}\left ( f_{U-Net}   \left ( concat \left (  \mathbb D\left (KP^{\hat{K}}_{comp}, \hat{KP}^{I_{l}}_{comp}  \right ), F_{coarse}^{I_{l}} \right )  \right )   \right ) \otimes \mathcal M^{I_{l}}_{sparse},   
\end{equation}
\begin{equation}
\mathcal M^{I_{l}}_{occlusion}= P_{2}\left ( f_{U-Net}   \left ( concat \left (  \mathbb D\left (KP^{\hat{K}}_{comp}, \hat{KP}^{I_{l}}_{comp}  \right ), F_{coarse}^{I_{l}} \right )  \right )   \right ),   
\end{equation} where $P_{1}\left ( \cdot \right )$ and $P_{2}\left ( \cdot \right )$ denote two different functions to predict a mask map and an occlusion map. $\mathbb D\left (\cdot \right )$ and $concat\left (\cdot \right )$ denote the heatmap difference operation and feature concatenation process.

\vspace{-0.3cm}
\subsection{Multi-task Learning-based Frame Generation \& Vertex Prediction}
\vspace{-0.2cm}
 
The proposed Sparse2Dense framework achieves seamless integration of frame generation and vertex prediction through a carefully designed multi-task learning architecture. In particular, the texture feature of the VVC reconstructed key-reference frame (i.e., $F_{texture}^{\hat{K}}$) serves as a pivotal bridge between the two tasks, enabling cross-task knowledge transfer while maintaining task-specific optimization pathways. In detail, a feature-level warping strategy is used to transform $F_{texture}^{\hat{K}}$ with the dense motion flow  $\mathcal M^{I_{l}}_{flow}$, followed by occlusion map $\mathcal M^{I_{l}}_{occlusion}$—guided refinement to enhance reconstruction fidelity.   
The refined feature $F_{refine}^{I_{l}}$ can be expressed as,
\begin{equation}
{F_{refine}^{I_{l}} =\mathcal M ^{I_{l}}_{occlusion} \odot  f_{w}\left (F_{texture}^{\hat{K}}, \mathcal M^{I_{l}}_{flow} \right )},
\end{equation} where $\odot $ is the Hadamard product to indicate the feature map regions with the corresponding confidences.

\textbf{Frame Generation.} Our framework leverages the GAN architecture as a generator due to their superior inference speed and deployment efficiency compared to other generative models. After obtaining the refined feature $F_{refine}^{I_{l}}$, the generator's network layers $G_{frame}\left ( \cdot \right )$ is used to convert $F_{refine}^{I_{l}}$ into human frames ${\hat{I}}_{l}$ as,  
\begin{equation}
{\hat{I}}_{l} =  G_{frame} \left (F_{refine}^{I_{l}}  \right ),
\end{equation}
where ${\hat{I}}_{l}$ is further fed to a discriminator to approximate the distribution of ${I}_{l}$.

\textbf{Vertex Prediction.} The goal of this vertex prediction module is to transform $F_{refine}^{I_{l}}$ into a geometric representation while maintaining computational efficiency. Herein, it operates in three stages: 1) Feature Enhancement:
$F_{refine}^{I_{l}}$ undergo spatial refinement through a sequence of residual blocks $ResBlocks \left (\cdot \right ) $ to preserve channel dimensions while enhancing semantic representations through skip connections.
2) Global Context Aggregation: The enhanced features are compressed via adaptive average pooling $AdaPool \left (\cdot \right ) $ to obtain global descriptors, capturing holistic spatial relationships. 3) Coordinate Regression: A fully connected layer $FC \left (\cdot \right ) $ maps the global descriptors to a flattened vector of size 20950, which is then reshaped into  structured 2D vertex coordinates $\mathbf{V} \in \mathbb R^{10475\times 2} $. It can be defined as, 
\begin{equation}
\mathbf{V}=Reshape\left ( FC \left (AdaPool\left ( ResBlocks\left ( F_{refine}^{I_{l}} \right )  \right )   \right )  \right ), 
\end{equation} where the formulation of $\mathbf{V}$ is supervised via the one-stage whole-body mesh recovery method (OSX)~\cite{lin2023one} during the training process.

\subsection{Model Optimization}

The unified framework integrates compression and vertex estimation, enabling joint end-to-end optimization with minimal computational overhead. The training objective is formulated by combining the following complementary losses: (1) equivariance loss $\mathcal L_{equ}$ and keypoint prior loss $\mathcal L_{kp}$ for keypoint converge and learning, (2) perceptual loss $\mathcal L_{per}$ and adversarial loss $\mathcal L_{adv}$ for high-fidelity signal synthesis, and (3) vertex loss $\mathcal L_{ver}$ for vertex prediction. 
\begin{equation}
\mathcal L_{total}= \lambda _{equ}\mathcal L _{equ}+\lambda _{kp}\mathcal L _{kp} 
+\lambda _{per}\mathcal L _{per}+\lambda _{adv}\mathcal L _{adv}
+\lambda _{ver}\mathcal L _{ver},
\end{equation} 
where $\lambda_{equ}$, $\lambda_{kp}$ and $\lambda_{per}$ are set to 10. $\lambda _{adv}$ and $\lambda _{ver}$ are set to 1 and 100, respectively.

\begin{figure}[t]
\centering
\vspace{-1cm}
\centerline{\includegraphics[width=1 \textwidth]{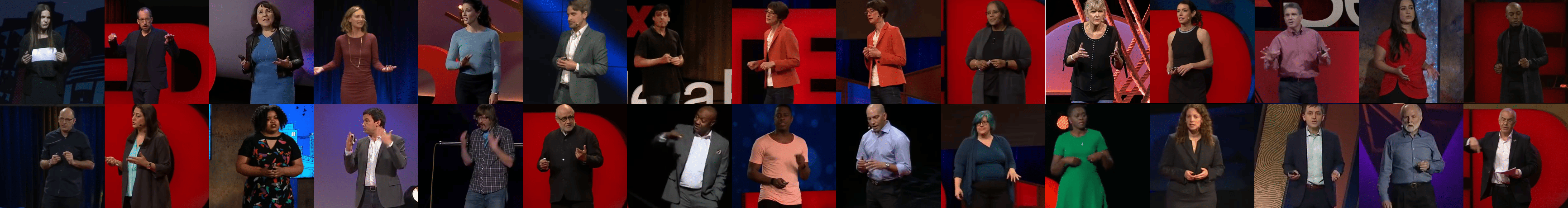}}
\vspace{-0.5cm}
\caption{Illustration of 30 test sequences selected and pre-processed from TED-Talk dataset~\cite{siarohin2021motion}.} 
\vspace{-1em}
\label{fig3}
\end{figure}

\section{Experimental Results}
This section begins with an introduction of the experimental settings, followed by a comprehensive analysis of the experimental results.
\subsection{Experimental Settings}
\qquad \textbf{Implementation Details.} The model is developed with PyTorch and trained on the TED-Talk dataset~\cite{siarohin2021motion}, comprising 1,132 training and 128 testing videos at 384×384 resolution. Training utilized 300 epochs with 30× repeat strategy across 4 NVIDIA TESLA V100 GPUs. Optimization employed Adam ($\beta _{1}$=0.5, $\beta _{2}$ =0.999) at a learning rate of 0.0002 to balance convergence stability and training efficiency.

\textbf{Compared Algorithms.} The proposed Sparse2Dense framework is benchmarked against VVC~\cite{bross2021overview} and eight generative human video compression methods (MRAA~\cite{siarohin2021motion}, FV2V~\cite{wang2021one}, TPSM~\cite{zhao2022thin}, CFTE~\cite{CHEN2022DCC}, LIA~\cite{wanglatent}, MTTF~\cite{yin2024generative}, IMT~\cite{chen2025rethinking}, IHVC~\cite{chen2025compressing}). All models are retrained on the TED-Talk dataset~\cite{siarohin2021motion} using identical parameters for fair comparisons. As shown Figure~\ref{fig3}, there are 30 selected test sequences (150 frames, 384×384 resolution). Evaluation follows JVET GFVC AhG test conditions~\cite{JVET-AJ2035}, with implementation details provided for each baseline as follows,
\begin{itemize}
\vspace{-1mm}
\item{\textbf{VVC Baseline:} VTM-22.2 (VVC reference software) serves as the baseline with LDB mode and QPs={37,42,47,52}. All raw videos are converted to YUV 4:2:0 format prior encoding.}
\vspace{-1.5mm}
\item{\textbf{Generative Codecs:} Key-reference frames undergo YUV 4:2:0 conversion and VTM-22.2 compression (QPs={22,32,42,52}). Inter-frames are encoded into compact parameters using context-adaptive arithmetic coding.}
\end{itemize}

\textbf{Evaluation Measures.} Traditional pixel-level metrics (e.g., PSNR/SSIM)~\cite{10477607,GFVC_Survey} may not be appropriate for generative video compression evaluation. We therefore utilize three learning-based perceptual metrics (DISTS~\cite{DISTS}, LPIPS~\cite{LPIPS}, FVD~\cite{unterthiner2019fvd}) for compressed human video quality assessment, where lower scores indicate superior perceptual quality. Figure~\ref{fig_RD} employs normalized 1-DISTS, 1-LPIPS, and 5000-FVD to ensure consistent rate-distortion curves. Compression efficiency is further quantified by encoded bitrates (kbps).

\subsection{Performance Comparisons}

\begin{figure*}[!t]
\centering
\vspace{-4em}
\subfloat[Rate-DISTS]{\includegraphics[width=0.333 \textwidth]{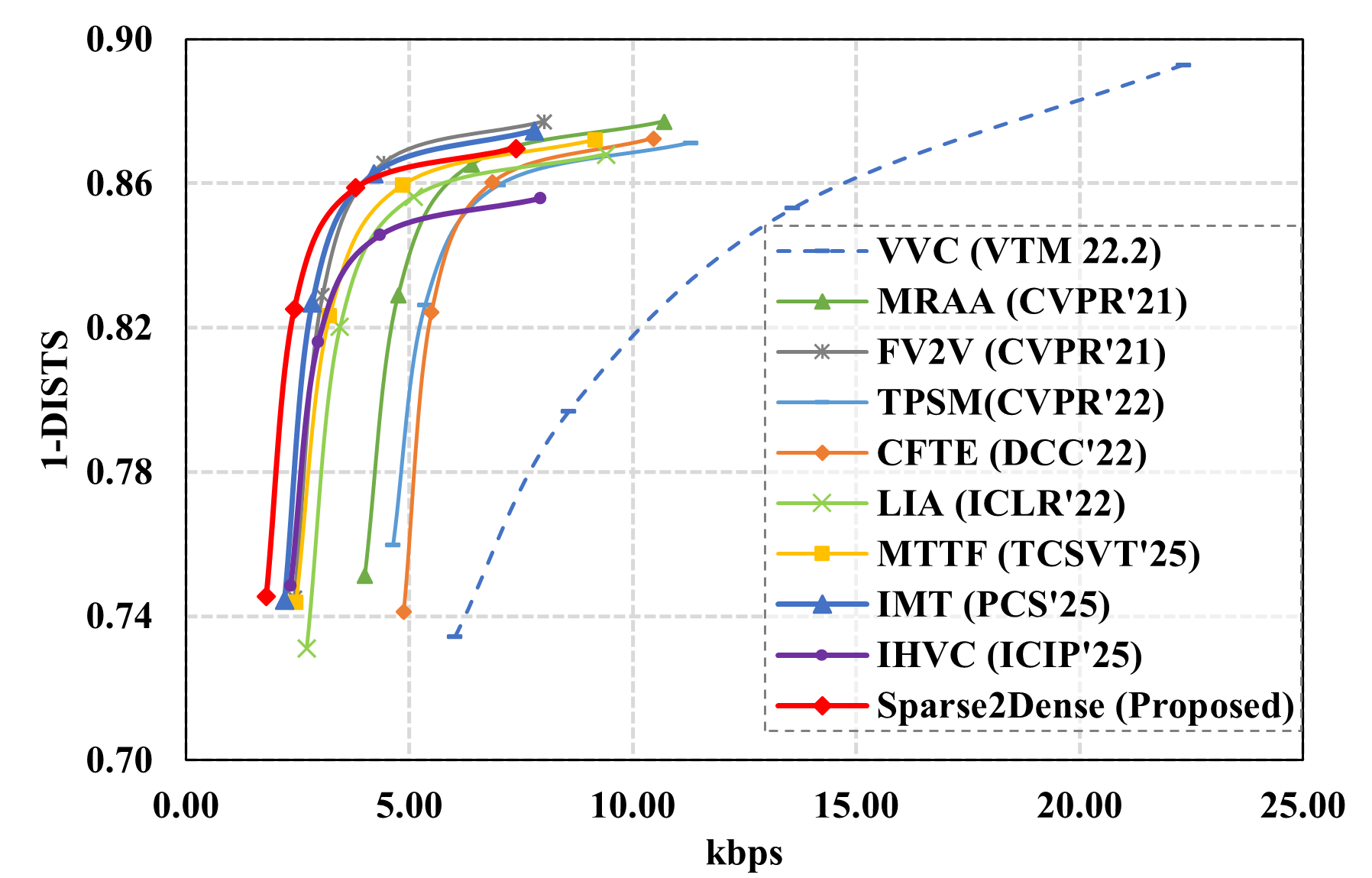}}
\subfloat[Rate-LPIPS]{\includegraphics[width=0.333 \textwidth]{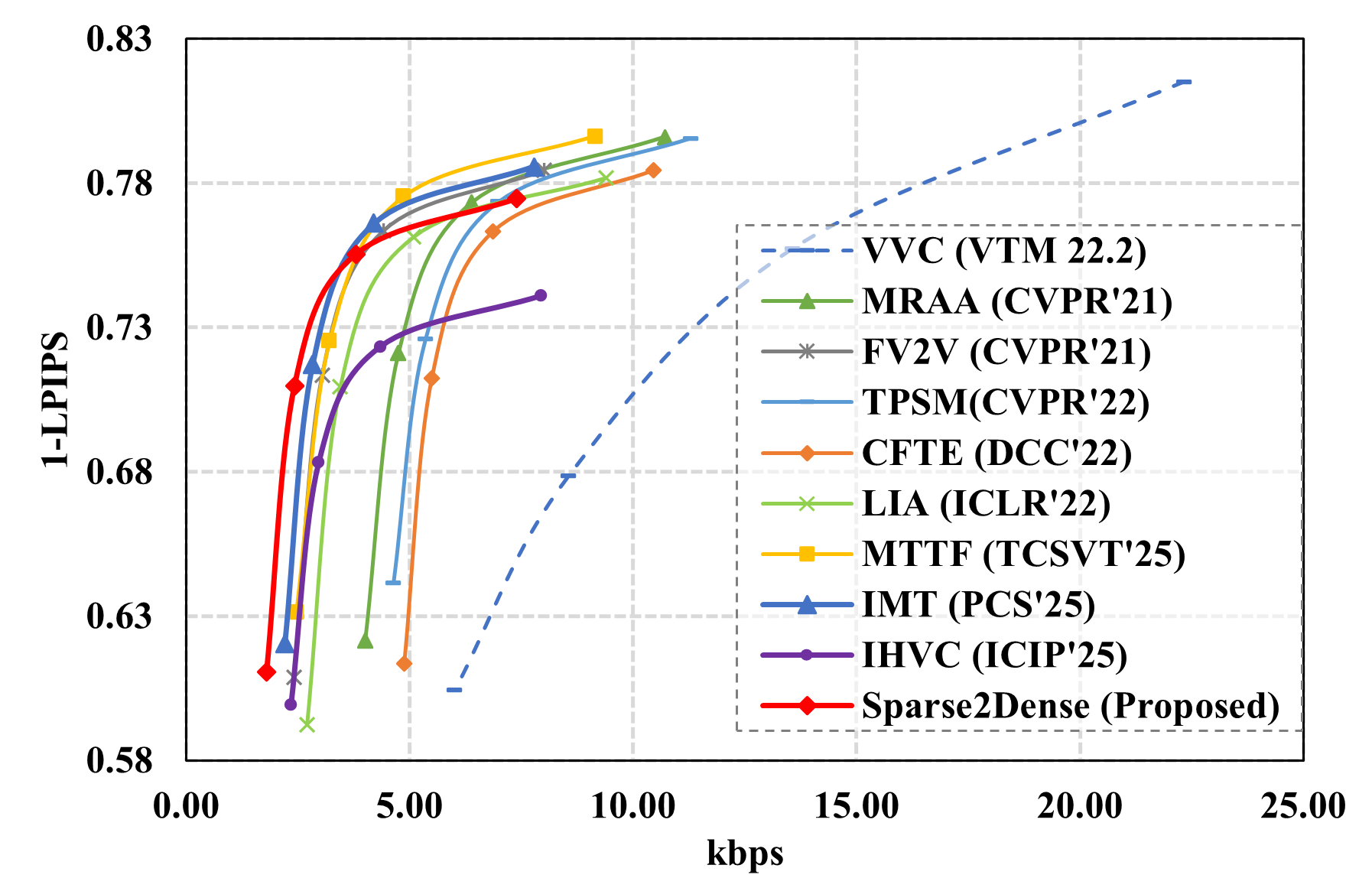}}
\subfloat[Rate-FVD]{\includegraphics[width=0.333 \textwidth]{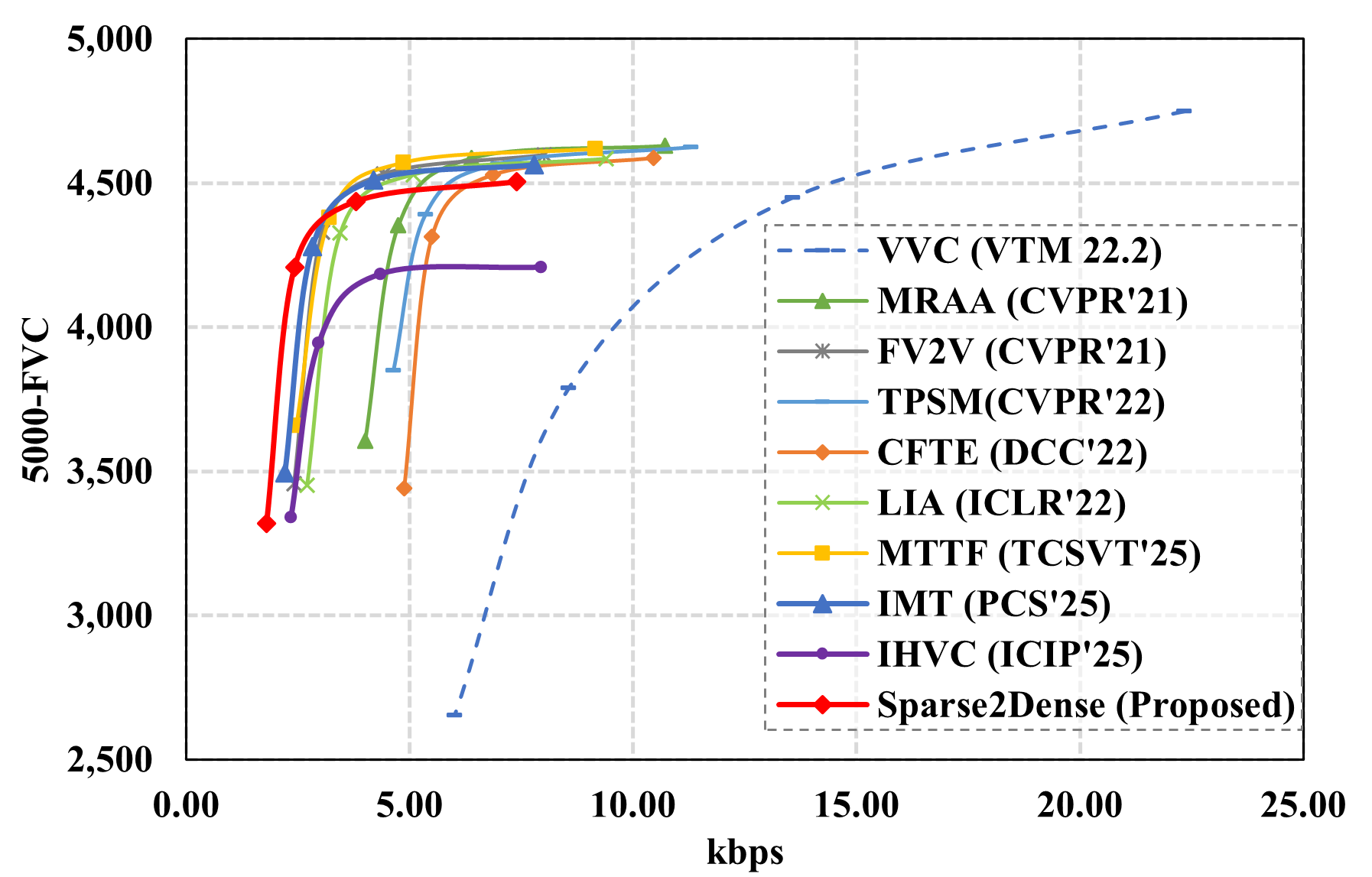}}
\vspace{-0.3cm}
\caption{Overall RD performance comparisons with VVC~\cite{bross2021overview}, MRAA~\cite{siarohin2021motion}, FV2V~\cite{wang2021one}, TPSM~\cite{zhao2022thin}, CFTE~\cite{CHEN2022DCC}, LIA~\cite{wanglatent}, MTTF~\cite{yin2024generative}, IMT~\cite{chen2025rethinking} and IHVC~\cite{chen2025compressing} in terms of rate-DISTS, rate-LPIPS and rate-FVD. }
\label{fig_RD}  
\vspace{-0.8em}
\end{figure*}

\begin{figure*}[t]
\centering
\subfloat[Similar Bitrate Comparisons]{\includegraphics[width=0.49 \textwidth]{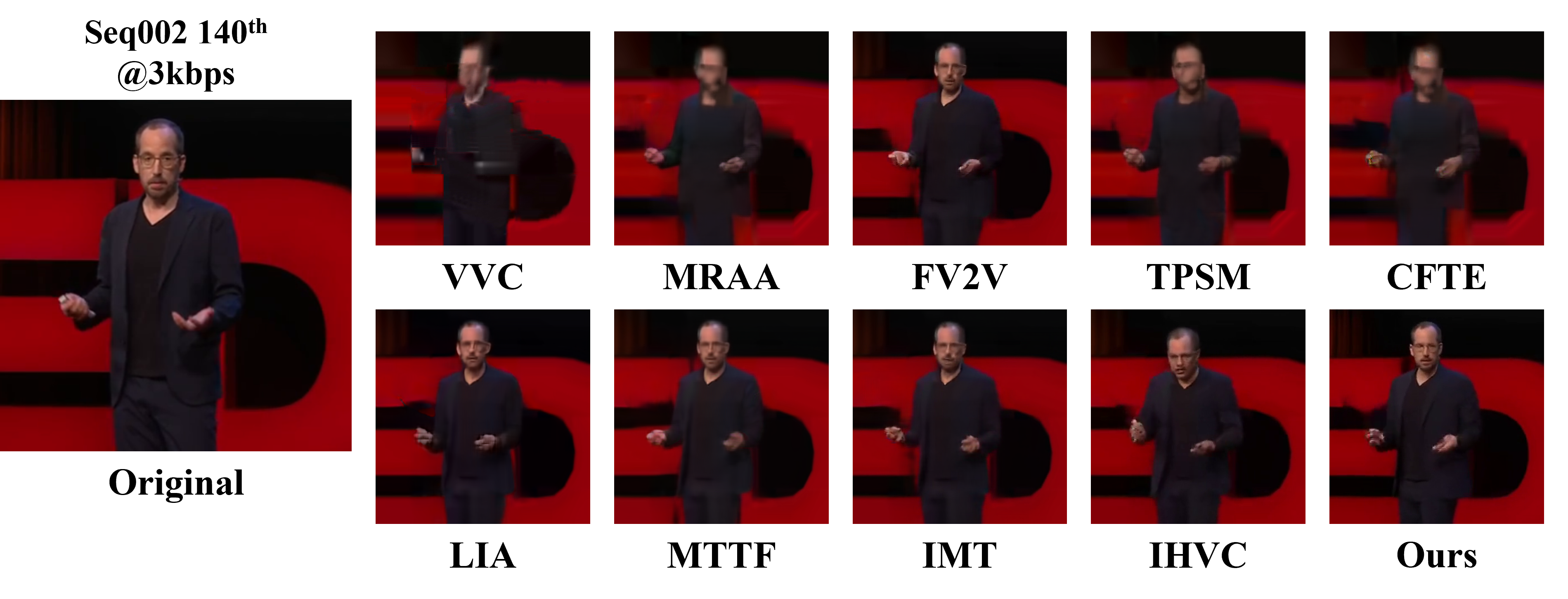}} 
\subfloat[Similar Quality Comparisons]{\includegraphics[width=0.49\textwidth]{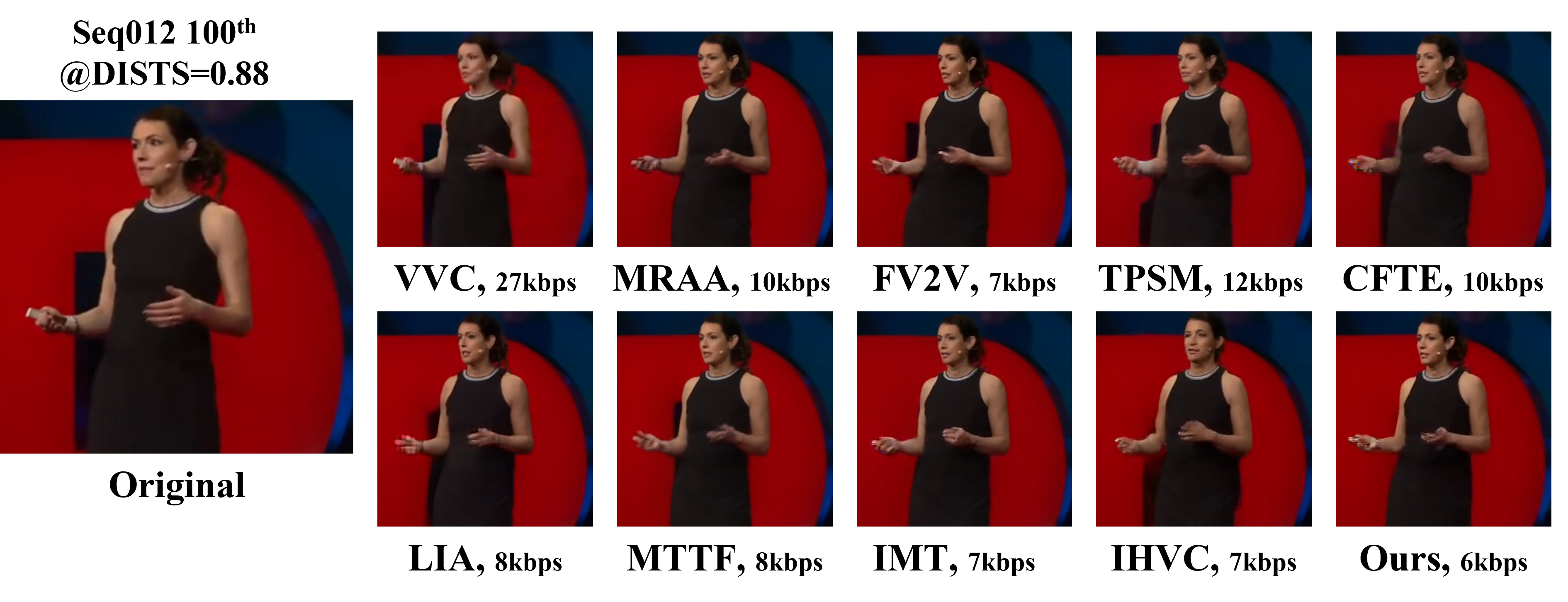}}
\vspace{-0.8em}
\caption{Visual quality comparisons among VVC~\cite{bross2021overview}, MRAA~\cite{siarohin2021motion}, FV2V~\cite{wang2021one}, TPSM~\cite{zhao2022thin}, CFTE~\cite{CHEN2022DCC}, LIA~\cite{wanglatent}, MTTF~\cite{yin2024generative}, IMT~\cite{chen2025rethinking} and IHVC~\cite{chen2025compressing} and Ours at the similar bitrate and quality. }
\label{fig_sub}  
\vspace{-1em}
\end{figure*}

\qquad \textbf{Rate-Distortion Performance.} Figure~\ref{fig_RD} compares the rate-distortion (RD) performance of Sparse2Dense with VVC and eight generative methods on TED-Talk test sequences using three perceptual metrics. Results indicate competitive RD efficiency against state-of-the-art VVC, especially in learning-based metrics, while achieving better performance than generative approaches through compact representation. However, at elevated bitrate ranges at 5$\sim $8kbps, our method exhibits performance limitations where objective quality slightly lags behind IMT and FV2V algorithms. Table \ref{table1} presents bit-rate savings of the proposed Sparse2Dense across 30 sequences, showing substantial improvements over VVC. Specifically, Sparse2Dense achieves 74.54\% (DISTS), 73.68\% (LPIPS), and 75.39\% (FVD) bit-rate reduction compared to VVC, surpassing all existing generative approaches.

\begin{table*}[t]
\renewcommand\arraystretch{1.2}
\caption{Average Bit-rate savings of 30 testing sequences in terms of Rate-DISTS, Rate-LPIPS and Rate-FVD }  
\label{table1}
\centering
\resizebox{1\textwidth}{!}{
\begin{tabular}{cccccccccc}
\hline
\begin{tabular}[c]{@{}c@{}}Anchor:VVC \\ (VTM 22.2)\end{tabular} & \begin{tabular}[c]{@{}c@{}}MRAA \\ (CVPR'21)\end{tabular} & \begin{tabular}[c]{@{}c@{}}FV2V\\  (CVPR'21)\end{tabular} & \begin{tabular}[c]{@{}c@{}}TPSM\\ (CVPR'22)\end{tabular} & \begin{tabular}[c]{@{}c@{}}CFTE \\ (DCC'22)\end{tabular} & \begin{tabular}[c]{@{}c@{}}LIA \\ (ICLR'22)\end{tabular} & \begin{tabular}[c]{@{}c@{}}MTTF\\  (TCSVT'25)\end{tabular} & \begin{tabular}[c]{@{}c@{}}IMT \\ (PCS'25)\end{tabular} & \begin{tabular}[c]{@{}c@{}}IHVC\\  (ICIP'25)\end{tabular} & \begin{tabular}[c]{@{}c@{}}\textbf{Sparse2Dense} \\ \textbf{(Propose)}\end{tabular} \\ \hline
Rate-DISTS                                                       & -50.92\%                                                  & -68.49\%                                                  & -44.51\%                                                 & -61.99\%                                                 & -40.33\%                                                 & -65.96\%                                                   & -70.52\%                                                & -66.68\%                                                  & \textbf{-74.54\%}                                                 \\
Rate-LPIPS                                                       & -51.02\%                                                  & -67.18\%                                                  & -47.22\%                                                 & -61.77\%                                                 & -40.51\%                                                 & -68.16\%                                                   & -70.56\%                                                & -62.65\%                                                  & \textbf{-73.68\%}                                                 \\
Rate-FVD                                                         & -55.66\%                                                  & -70.69\%                                                  & -52.08\%                                                 & -66.94\%                                                 & -44.73\%                                                 & -71.27\%                                                   & -72.46\%                                                & -64.82\%                                                  & \textbf{-75.39\%}                                                 \\ \hline
\end{tabular}
}
\vspace{-1em}
\end{table*}

\begin{figure*}[t]
\centering
\subfloat[Vertex Prediction from 3D Keypoints]{\includegraphics[width=0.47 \textwidth]{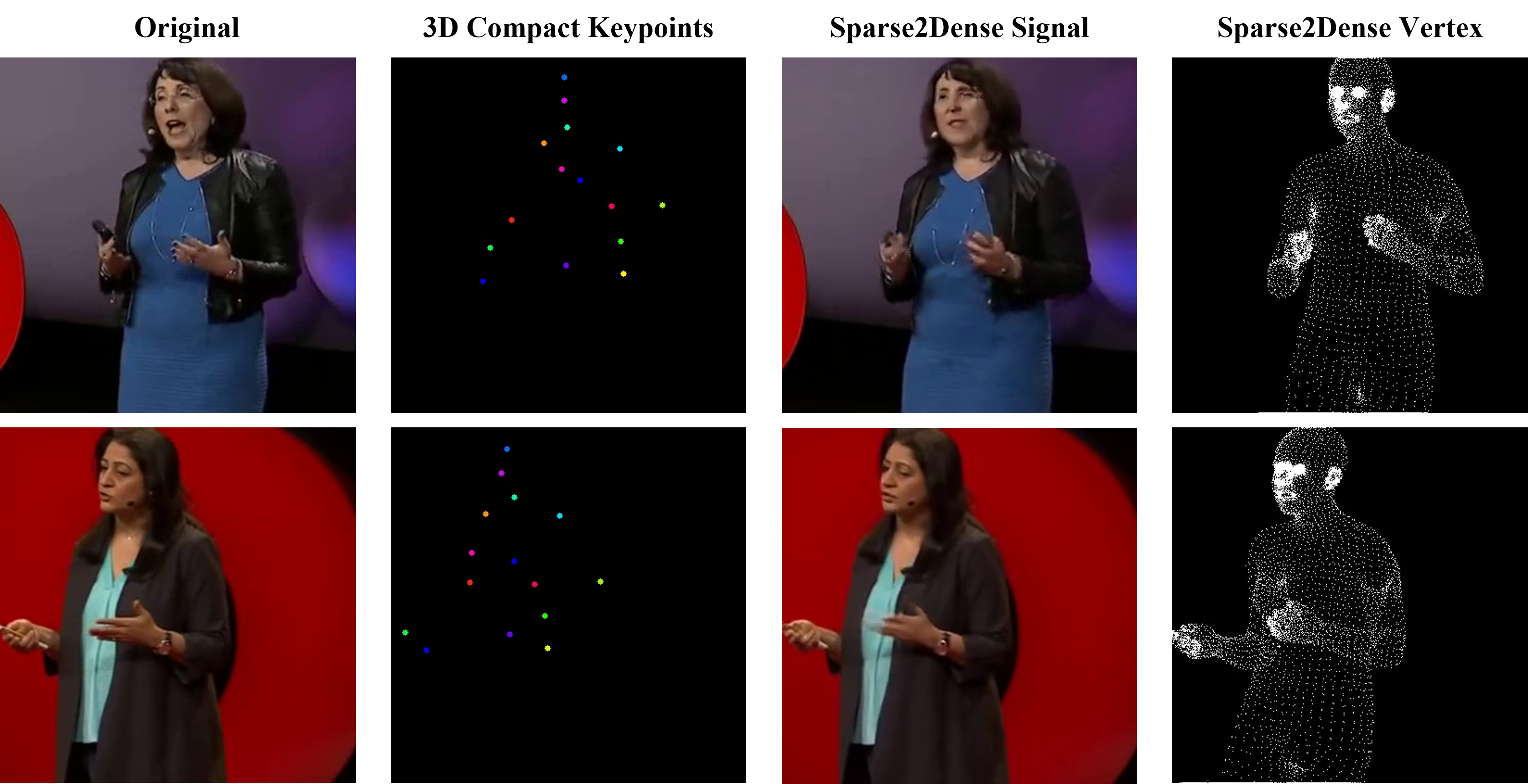}} 
\hspace{0.5cm}
\subfloat[Vertex Prediction Error Comparisons]{\includegraphics[width=0.47\textwidth]{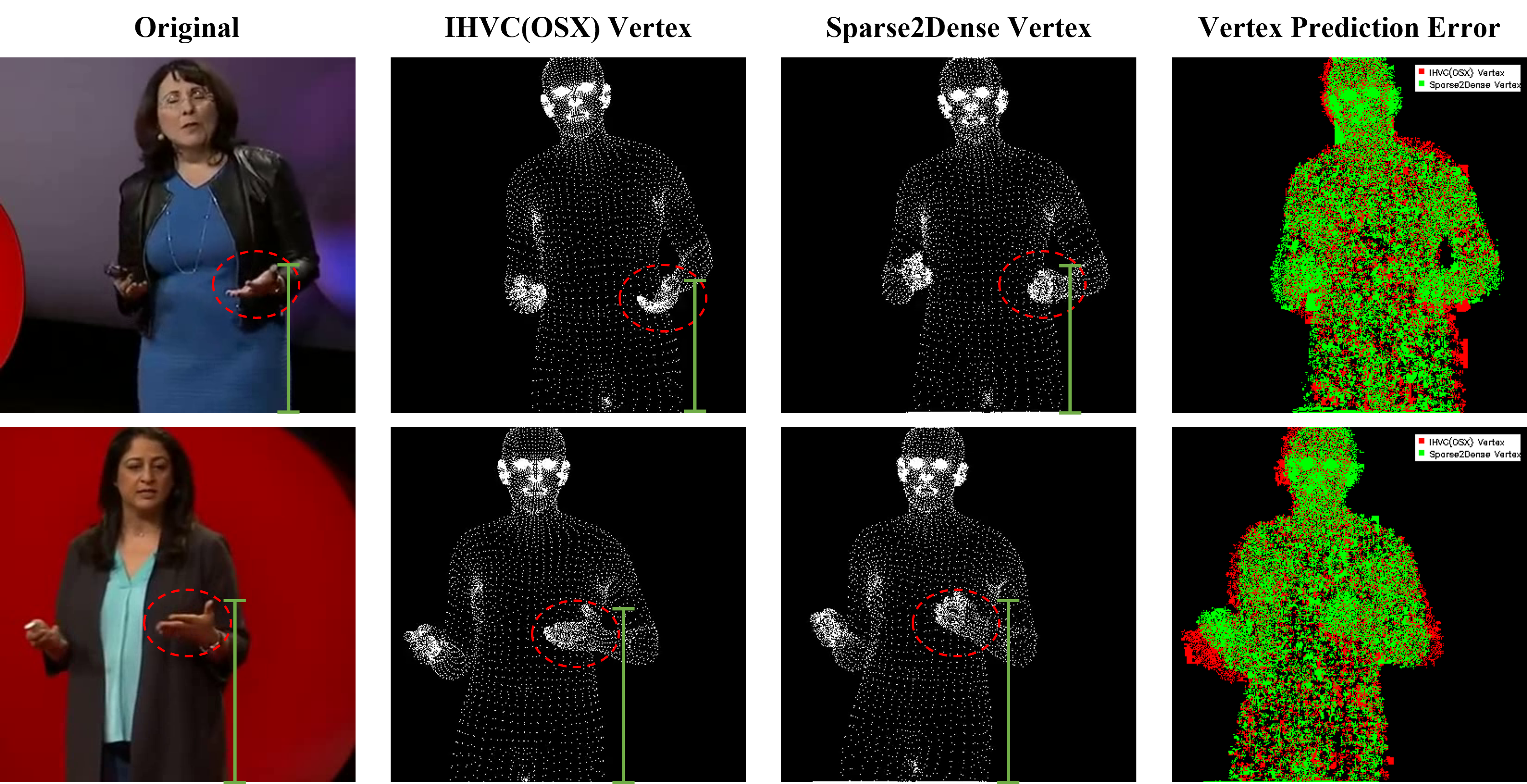}}
\vspace{-1.2em}
\caption{Evaluation of vertex prediction performance using the proposed Sparse2Dense.}
\label{vertex}  
\vspace{-2em}
\end{figure*}

\textbf{Subjective Quality Comparisons.} Fig. \ref{fig_sub} shows visual comparisons under similar bitrate and quality conditions across compression methods. Sparse2Dense outperforms VVC in reconstructing human videos, producing fewer artifacts while maintaining higher visual quality. When compared with generative human video coding algorithms, Sparse2Dense preserves finer textures and suppresses background noise more effectively, resulting in perceptually coherent reconstructions with better detail perseverations.

\textbf{Vertex Prediction Results.} As shown in Figure~\ref{vertex}, the proposed Sparse2Dense compresses high-dimensional signals into 3D compact keypoints for efficient transmission, enabling accurate reconstruction of vertices and 2D visual signals at the decoder. Since our method uses the OSX model's vertex output as supervision during training, we adopt its predictions as the compared ground truth. While IHVC (OSX) preserves plausible hand poses but suffers from positional deviations, our approach achieves more accurate vertex localization at the cost of less precise pose recovery. This highlights both the benefit of multi-task learning in improving vertex location prediction and the room for further accuracy enhancement. Moreover, our method is significantly faster—0.002s vs. 0.724s per frame—making it highly suitable for real-time processing and prediction applications.


\section{Conclusion}
This work addresses the critical challenge of achieving ultra-low bitrate human video compression while maintaining accurate vertex prediction in bandwidth-constrained multimedia applications. Specifically, we propose a keypoint-driven generative framework, namely Sparse2Dense,  which encodes complex human motion into compact keypoint representations at the encoder side and produce high-fidelity human video as well as accurate vertex information at the decoder side. Notably, a learnable vertex predictor is introduced into keypoint-aware deep generative model to jointly optimize geometric vertex structures with video generation, ensuring alignment between visual reconstruction and geometry prediction. Experimental results demonstrate that Sparse2Dense outperforms traditional codecs and generative video compression approaches in compression efficiency, reconstruction quality and vertex prediction accuracy, making it promising for versatile human-centric media communications.

\vspace{-0.7em}
\section*{References}
\vspace{-0.6em}
\bibliographystyle{IEEEtran}
\bibliography{main}
\vspace{-0.8em}
\end{document}